\documentclass[a4paper, 11pt, oneside]{article} % A4 paper size, default 11pt font size and oneside for equal margins

\usepackage{geometry}
\usepackage{booktabs} % For formal tables

\usepackage{mathptmx} % This is Times font

\usepackage{fancyhdr}
\usepackage[normalem]{ulem}
\usepackage{microtype}
\usepackage{graphicx}
\usepackage{subfigure}
\usepackage{colortbl}
\definecolor{mygray}{gray}{.88}
\usepackage{multirow}
\usepackage{amsmath}
\usepackage{bm}
\usepackage{epsfig}
\usepackage{authblk}

\usepackage[hyphens]{url}
\usepackage{hyperref}
\usepackage{color}
\usepackage{soul}
\usepackage{bbding}
\definecolor{mygray}{gray}{.88}

\newcommand{\tabincell}[2]{\begin{tabular}{@{}#1@{}}#2\end{tabular}}

\begin{document} 

%%%%封面内容编辑%%%%
\begin{titlepage} % Suppresses headers and footers on the title page

	\centering % Centre everything on the title page
	
	\scshape % Use small caps for all text on the title page
	
	\vspace*{\baselineskip} % White space at the top of the page
	
	%------------------------------------------------
	%	Title
	%------------------------------------------------
	
	\rule{\textwidth}{1.6pt}\vspace*{-\baselineskip}\vspace*{2pt} % Thick horizontal rule
	\rule{\textwidth}{0.4pt} % Thin horizontal rule
	
	\vspace{0.75\baselineskip} % Whitespace above the title
	
	{\LARGE AIBench: \\An Industry Standard Internet Service\\ AI Benchmark Suite\\} % Title
	
	\vspace{0.75\baselineskip} % Whitespace below the title
	
	\rule{\textwidth}{0.4pt}\vspace*{-\baselineskip}\vspace{3.2pt} % Thin horizontal rule
	\rule{\textwidth}{1.6pt} % Thick horizontal rule
	
	\vspace{2\baselineskip} % Whitespace after the title block
	
	%------------------------------------------------
	%	Subtitle
	%------------------------------------------------
	
	%Subtitle here % Subtitle or further description
	
	\vspace*{3\baselineskip} % Whitespace under the subtitle
	
	%------------------------------------------------
	%	Editor(s)
	%------------------------------------------------
	
	Edited By
	
	\vspace{0.5\baselineskip} % Whitespace before the editors
	
	{\scshape\Large Wanling Gao\\ Fei Tang\\ Lei Wang\\Jianfeng Zhan\\Chunxin Lan\\ Chunjie Luo\\Yunyou Huang\\Chen Zheng\\Jiahui Dai\\Zheng Cao\\et al.\\ }
	
	%{\scshape\Large Wanling Gao\\ Fei Tang\\ Lei Wang\\Jianfeng Zhan\\Chunxin Lan\\ Chunjie Luo\\Yunyou Huang\\Jiahui Dai\\Hainan Ye\\Zheng Cao\\Daoyi Zheng\\Haoning Tang\\Kent Zhan\\Biao Wang\\Defei Kong\\Shimin Gong\\Minghe Yu\\Chongkang Tan\\Yabin Huang\\Xinhui Tian\\Yatao Li\\Junchao Shao\\Xiaoyu Wang\\Zhenyu Wang\\ } % Editor list
	
	\vspace{0.5\baselineskip} % Whitespace below the editor list

	\vfill % Whitespace between editor names and publisher logo
	
	%------------------------------------------------
	%	Publisher
	%------------------------------------------------
	
	%\plogo % Publisher logo
	%\def\BUlogo{\epsfig{file=ICT.pdf,height=3cm}}
	%\includegraphics[scale=0.135]{ICT.pdf}
	\epsfig{file=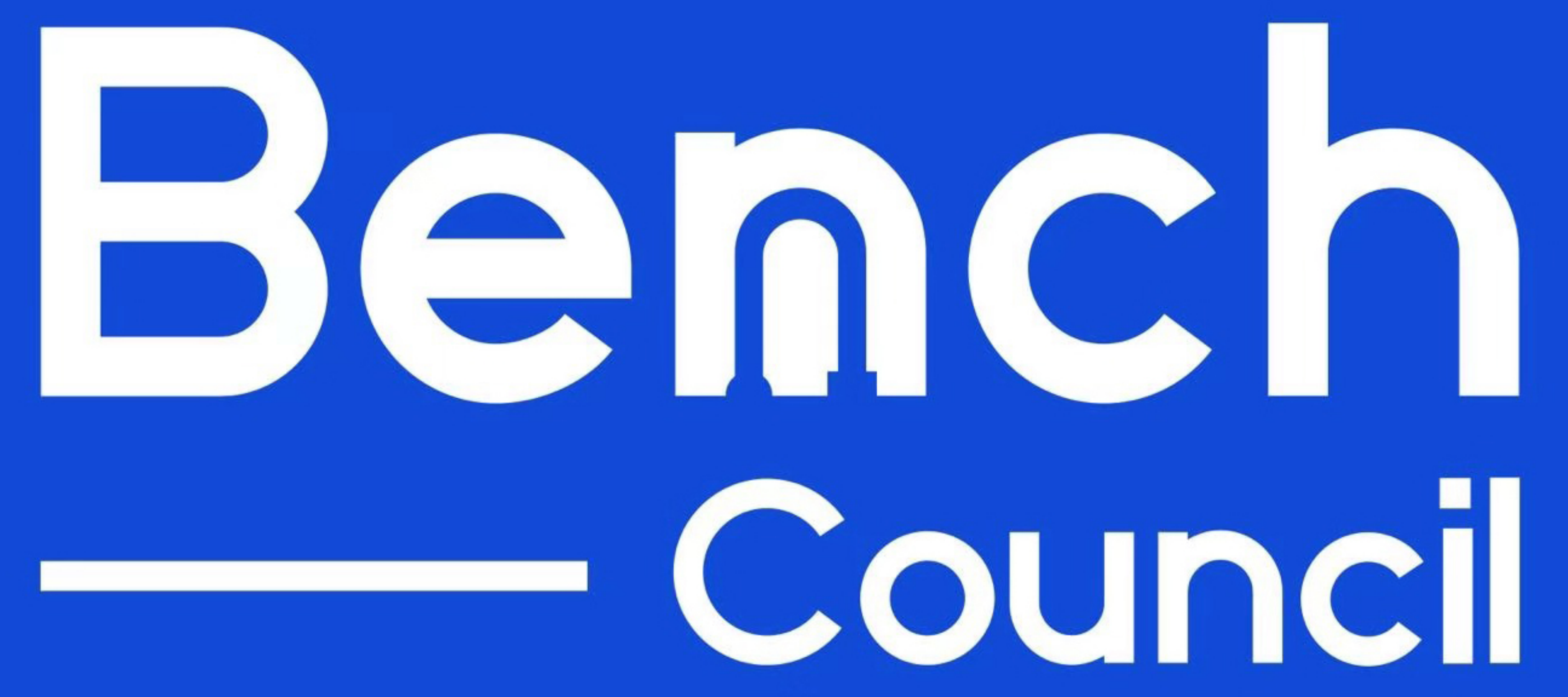,height=2cm}
	\textit{\\BenchCouncil: International Open Benchmarking Council\\Chinese Academy of Sciences\\Beijing, China\\http://www.benchcouncil.org/AIBench/index.html} % Editor affiliation
	\vspace{5\baselineskip} % Whitespace under the publisher logo

	Technical Report No. BenchCouncil-AIBench-2019 % Publication year
	
	{\large Aug 12, 2019} % Publisher

\end{titlepage}

%----------------------------------------------------------------------------------------

%%%title here%%%
\title{AIBench: An Industry Standard Internet Service AI Benchmark Suite}

\author[1,2,4]{Wanling Gao}
\author[1,4]{Fei Tang\thanks{Wanling Gao and Fei Tang contributed equally to this paper.}}
\author[1,2,4]{Lei Wang}
\author[1,2,4]{Jianfeng Zhan\thanks{Jianfeng Zhan is the corresponding author.}}
\author[1]{Chunxin Lan}
\author[1,2,4]{Chunjie Luo}
\author[1,4]{Yunyou Huang}
\author[1,2,4]{Chen Zheng}
\author[3]{Jiahui Dai}
\author[6]{Zheng Cao}
\author[7]{Daoyi Zheng}
\author[8]{Haoning Tang}
\author[9]{Kunlin Zhan}
\author[10]{Biao Wang}
\author[11]{Defei Kong}
\author[12]{Tong Wu}
\author[13]{Minghe Yu}
\author[14]{Chongkang Tan}
\author[15]{Huan Li}
\author[16]{Xinhui Tian}
\author[17]{Yatao Li}
\author[18]{Gang Lu}
\author[19]{Junchao Shao}
\author[20]{Zhenyu Wang}
\author[21]{Xiaoyu Wang}
\author[3,5]{Hainan Ye}

\affil[1]{State Key Laboratory of Computer Architecture, Institute of Computing Technology, Chinese Academy of Sciences \\ \{gaowanling, tangfei, wanglei\_2011, zhanjianfeng, lanchuanxin\}@ict.ac.cn}
\affil[2]{BenchCouncil (International Open Benchmarking Council)}
\affil[3]{Beijing Academy of Frontier Sciences and Technology, \{daijiahui,yehainan\}@mail.bafst.com}
\affil[4]{University of Chinese Academy of Sciences}
\affil[5]{Xinxiu (SciCom)}
\affil[6]{Alibaba, zhengzhi.cz@alibaba-inc.com}
\affil[7]{Baidu, zhengdaoyi@baidu.com}
\affil[8]{Tencent, haoningtang@tencent.com}
\affil[9]{58.com, zhankunlin@58.com}
\affil[10]{NetEase, bjwangbiao@corp.netease.com}
\affil[11]{ByteDance, kongdefei@bytedance.com}
\affil[12]{China National Institute of Metrology, wut@nim.ac.cn}
\affil[13]{Zhihu, yuminghe@zhihu.com}
\affil[14]{Lenovo, tanck1@lenovo.com}
\affil[15]{Paypal, huanli1@paypal.com}
\affil[16]{Moqi, xinhuit@moqi.ai}
\affil[17]{Microsoft Research Asia, yatli@microsoft.com}
\affil[18]{Huawei, lugang3@huawei.com}
\affil[19]{JD.com, shaojunchao@imdada.cn}
\affil[20]{CloudTa, wangzhenyu@cloudta.com.cn}
\affil[21]{Intellifusion, wang.xiaoyu@intellif.com}

\date{Aug 12, 2019}
\maketitle

\begin{abstract}

%Internet service providers, e.g., Google, Facebook, Alibaba, deploy datacenters to deliver good levels of service performance.
Today's Internet Services  are undergoing fundamental changes and shifting to an intelligent computing era where AI is widely employed to augment services. 
% performance AI features prominently and is deeply integrated into datacenter computing. 
In this context, many innovative AI algorithms, systems, and architectures are proposed, and thus the importance of benchmarking and evaluating them rises. 
However, modern Internet services adopt a microservice-based architecture and consist of various modules. The  diversity of these modules and complexity of execution paths, the massive scale and complex hierarchy of datacenter infrastructure, the confidential issues of data sets and workloads  pose great challenges to benchmarking. %An end-to-end industry benchmark that captures primary modules and critical path is demanded for the entire lifecycle performance evaluation of datacenters.
%the massive scale and complex hierarchy of datacenter infrastructure pose great challenges and pressures on datacenter AI benchmarking. 
%Previous work only collects a set of micro or component AI benchmarks and fails to depict the entire lifcycle of internet service. 

In this paper, we present the first industry-standard Internet service AI benchmark suite---AIBench with  seventeen industry partners, including several top Internet service providers.  AIBench provides a highly extensible, configurable, and flexible benchmark framework that contains loosely coupled modules.  %like data input, prominent AI problem domain, online inference (i.e., AI-as-a-service), offline training, and automatic deployment tool modules. 
We identify sixteen prominent AI problem domains like  learning to rank, each of which forms an AI component benchmark, from three most important Internet service domains: search engine, social network, and e-commerce, which is by far the most comprehensive AI benchmarking effort. % We abstract twelve fundamental units of computation accross different component benchmarks as the micro benchmarks. 
% ,  Collectively, AIBench can constitute diverse end-to-end industry benchmarks with underlying application models. 
%Individually, each AI component within the modules form a micro or component AI benchmark. 
On the basis of the AIBench framework, abstracting the real-world data sets and workloads from one of the top e-commerce providers, we design and implement the first end-to-end Internet service AI benchmark, which contains the primary modules in the critical paths of an industry scale application and is scalable to deploy on different cluster scales. The preliminary evaluation has shown the value of our benchmark suite  with respect to the previous documented performance models and insights without the publicly available ensemble of Internet service data sets, workloads and user logs.  The specifications, source code, and performance numbers are publicly available from the benchmark council web site \url{http://www.benchcouncil.org/AIBench/index.html}.
% that are not only AI-intensive but also containing the modules in the critical path. 
%For fine-grained evaluation, sixteen component benchmarks and twelve micro benchmarks are also included.

\end{abstract}

\clearpage

\section{Introduction}

Modern Internet service providers pervasively employ AI to augment their services, as the advancement of AI technology has brought breakthroughs in processing images, video, speech, and audio ~\cite{lecun2015deep}, 
%Gartner analysts report that AI is one of the top ten trends that most impact the infrastructure~\cite{gartner2018}, and predicts that AI will be introduced in almost every products or service by 2020~\cite{gartner2017}. Benefiting from AI technology, many Internet service companies have made significant strides towards improving serving efficiency, 
and hence boost the deployments of massive AI algorithms, systems and architectures. For example, Alibaba proposes a new DUPN network for more effective personalization~\cite{ni2018perceive}. Facebook integrates AI into many essential products and services like news feed~\cite{hazelwood2018applied}. Google proposes the TensorFlow~\cite{abadi2016tensorflow} system and the tensor processing unit (TPU)~\cite{jouppi2017datacenter} to accelerate the service performance. Amazon adopts AI for intelligent product recommendation~\cite{smith2017two}. 

Consequently, the pressures of measuring and evaluating these algorithms, systems, and architectures rise.
%On one hand
First, the real world data sets and workloads from Internet services are treated as first-class  confidential issues by their providers, and  they  are isolated between academia and industry, or even among different  providers. However,  there are only a few publicly available performance model or observed insights~\cite{hazelwood2018applied,ayers2018memory} about industry-scale Internet services that can be leveraged for further research. %as research collaboration work from both industry and academia. %from Internet services providers. 
As there is no publicly available industry-scale Internet service benchmark, the state-of-the-art and state-of-the-practice are advanced only by the research staffs inside Internet service providers, which is not sustainable and poses a huge obstacle for our communities towards developing an open and mature research field. 

Second, AI has infiltrated into almost all aspects of Internet services, ranging from offline analytics to online service. Thus, 
to cover the critical paths and characterize prominent characteristics of a realistic AI scenairo, end-to-end application benchmarks should be provided~\cite{barroso2009datacenter,bigbench}.
Meanwhile, there are many classes of Internet services. Modern Internet services workloads expand and change very fast, and it is not scalable or even impossible 
to create a new benchmark or proxy for every possible workload~\cite{gao2018motif}.  Moreover, data sets have great impacts on system and microarchitectural characteristics~\cite{xie2018cvr,gao2018motif}, so diverse data inputs should be considered. So we need identify representative data sets, abstract the prominent AI problem domains (component benchmarks), and further understand what are the most intensive  units of computation (micro benchmarks), on the basis of which, we can build a concise and comprehensive AI benchmark framework. 
%an datacenter AI benchmark suite must be an end-to-end application~\cite{barroso2009datacenter,bigbench}. 

Finally but not least, from an architectural perspective, porting a full-scale AI applications to a new architecture at an earlier stage is difficult and even impossible~\cite{bailey1991parallel}, while using micro or component benchmarks alone are insufficient to discover the time breakdown of different modules and locate the bottleneck within a realistic AI application scenario at a later stage~\cite{bailey1991parallel}. 
Hence, a realistic  AI benchmark suite should have the ability to run not only collectively as a whole end-to-end application to discover the  time breakdown of different modules but also individually as a micro or component benchmark for fine tuning hot spot functions or kernels. 
The state-of-the-art or state-of-the-practise ~\cite{mlperf,adolf2016fathom,deepbench,dong2017dnnmark,
coleman2017dawnbench} AI benchmarks only provide a few micro or component benchmarks, and none of them is able to cover the full use cases  of an  industry-scale  Internet service. So an industry standard Internet service AI benchmark suite consisting of a full spectrum of micro or component benchmarks and an end-to-end application benchmark is of great significance to bridge this huge gap.  %industry standard benchmark must be   there is no pulicly available end-to-end application benchmark. Instead,

To the best of our knowledge, this paper presents the first industry scale AI benchmark suite, AIBench, joint with seventeen industry partners. 
First, we present a highly extensible, configurable, and flexible benchmark framework, containing multiple loosely coupled modules like data input, prominent AI problem domains, online inference, offline training and automatic deployment tool modules. We analyze  typical AI application scenarios from three most important Internet services domains, including search engine, social network, and e-commerce, and then we abstract and identify sixteen prominent  AI problem domains, including \emph{ classification, image generation, text-to-text translation, image-to-text, image-to-image, speech-to-text, face embedding, 3D face recognition, object detection, video prediction, image compression, recommendation, 3D object reconstruction, text summarization, spatial transformer, and learning to rank}. We implement sixteen component benchmarks for those AI problem domains, and further profile and implement  twelve  fundamental units of computation across different component benchmarks as the micro benchmarks. 
%In a collective way, AIBench has the ability to constitute an end-to-end application benchmark with an underlying industry model, while in an individual way, each component within a module form a single micro or component benchmark for fine-grained evaluation.
On the basis of the AIBench framework, we design and implement the first end-to-end Internet service AI benchmark with an underlying e-commerce searching business model. As a whole, it covers the major modules and critical paths of an industry scale e-commerce provider. The application benchmark reuses ten component benchmarks from the AIBench framework, receives the query requests and performs
personalized searching, recommendation and advertising, integrated with AI inference and training. %To support scalable deployment on varying-scale datacenters, each module can be deployed on multiple nodes,
%To bridge the workload and data gaps between industries and academia, the  critical path and major modules are abstracted from real scenarios of our industry partner, meanwhile, 
The data maintains the real-world data characteristics through anonymization. Data generators are also provided to generate specified data scale, using several configurable parameters. 

In summary, our contributions are four-fold as follows.
%In summary, we make the following contributions in this paper.\vspace{2ex}

\begin{itemize}

\item We propose and implement a highly extensible, configurable, and flexible AI benchmark framework. %---supporting the construction of multiple end-to-end application benchmarks in a collective way, meanwhile, supporting the evaluation of single component in an individual way.
%The framework provides various AI algorithms targeting at the above problem domains, and is flexible and configurable to constitute a realistic end-to-end application. Meanwhile, each AI algorithm form an individual component or micro benchmark for fine-grained evaluation.
\item We identify sixteen prominent AI problem domains with seventeen industry partners, and implement sixteen component benchmarks targeting those domains accordingly. % implement twelve micro benchmarks and
\item We design and implement the first industry scale end-to-end  Internet service AI  benchmark suite, with an underlying e-commerce searching model.
%  On the basis of our AI benchmark framework, Moreover, sixteen component benchmarks and twelve micro benchmarks are provided individually. The framework is extensible to suit for other application benchmark construction like intelligent medicine.

\item On the CPU and GPU clusters, we perform workload characterizations of the end-to-end Internet service AI benchmark. We found that AI-related components significantly change the critical paths of the online service and deteriorate the average and tail latency, and the architects must perform a serious trade-off between the service quality and the complexity of neural network model. For offline training, we identify six kernels, consistent with our micro benchmarks~\footnote{The six kernels are named based on CUDA functions and they  are a subset of our micro benchmarks.}, and the corresponding function calls that takes up the most percentages of  running time, and analyze the GPU execution stalls that influence the performance, which provide the guide for further optimizations. 

\end{itemize}

The rest of this paper is organized as follows. In Section 2, we present the related work. Section 3 summarizes the AIBench framework. Section 4 illustrates the design and implementation of an end-to-end application benchmark. Section 5 illustrates the experiment configurations. In Section 6, we present the characterization results on GPUs and CPUs.
Finally, we draw the conclusion in Section 7.

\section{Related Work}

AI attracts great attention, appealing many research efforts on benchmarks. Table~\ref{comparition_table} compares AIBench  with respect to the state-of-the-art or state-of-the-practice AI benchmark suites, from the perspectives of end-to-end application benchmarks, component benchmarks, micro benchmarks, real-world data sets and software stacks.

\begin{table}[htbp]
\renewcommand\arraystretch{1.2}
\scriptsize
\centering
\caption{AI Benchmark Comparison.}\label{comparition_table}
\center
\begin{tabular}{|p{0.8in}|p{0.18in}|p{0.4in}|p{0.4in}|p{0.4in}|p{0.5in}|p{0.5in}|p{0.55in}|p{0.4in}|}
%\begin{tabular}{|p{0.9in}|p{0.63in}|p{0.58in}|p{0.55in}|p{0.55in}|p{0.55in}|p{0.4in}|p{0.42in}|}
\hline
\multicolumn{2}{|c|}{} & \tabincell{c}{AIBench} &\tabincell{c}{MLPerf} &\tabincell{c}{Fathom}
&\tabincell{c}{DeepBench}
&\tabincell{c}{DNNMark}
&\tabincell{c}{DAWNBench}
&\tabincell{c}{TBD}\\
\hline
\rowcolor{mygray} \multicolumn{9}{|l|}{End-to-End Application Benchmark}\\
\hline
\multicolumn{2}{|c|}{Online module} & \CheckmarkBold & $\times$ & $\times$ & $\times$ & $\times$ & $\times$ & $\times$\\
\hline
\multicolumn{2}{|c|}{Offline module} & \CheckmarkBold & $\times$ & $\times$ & $\times$ & $\times$ & $\times$ & $\times$\\
\hline
\rowcolor{mygray} \multicolumn{9}{|l|}{Component Benchmark}\\
\hline
\multirow{2}{*}{\tabincell{l}{Image\\classification}} & \tabincell{l}{Train} & \CheckmarkBold & \CheckmarkBold & \CheckmarkBold & $\times$ & $\times$ & \CheckmarkBold & \CheckmarkBold \\
\cline{2-9}
& \tabincell{l}{Infer} & \CheckmarkBold & \CheckmarkBold & \CheckmarkBold & $\times$ & $\times$ & \CheckmarkBold &  $\times$ \\
\hline
\multirow{2}{*}{\tabincell{l}{Image\\generation}} & \tabincell{l}{Train} & \CheckmarkBold & $\times$ & $\times$ & $\times$ & $\times$ & $\times$ & \CheckmarkBold \\
\cline{2-9}
& \tabincell{l}{Infer} & \CheckmarkBold & $\times$ & $\times$ & $\times$ & $\times$ & $\times$ & $\times$\\
\hline
\multirow{2}{*}{\tabincell{l}{Text-to-Text}} & \tabincell{l}{Train} & \CheckmarkBold & \CheckmarkBold & \CheckmarkBold & $\times$ & $\times$ & $\times$ & \CheckmarkBold\\
\cline{2-9}
& \tabincell{l}{Infer} & \CheckmarkBold & \CheckmarkBold & \CheckmarkBold & $\times$ & $\times$ & $\times$ & $\times$\\
\hline
\multirow{2}{*}{\tabincell{l}{Image-to-Text}} & \tabincell{l}{Train} & \CheckmarkBold & $\times$ & $\times$ & $\times$ & $\times$ & $\times$ & $\times$\\
\cline{2-9}
& \tabincell{l}{Infer} & \CheckmarkBold & $\times$ & $\times$ & $\times$ & $\times$ & $\times$ & $\times$\\
\hline
\multirow{2}{*}{\tabincell{l}{Image-to-Image}} & \tabincell{l}{Train} & \CheckmarkBold & $\times$ & $\times$ & $\times$ & $\times$ & $\times$ & $\times$\\
\cline{2-9}
& \tabincell{l}{Infer} & \CheckmarkBold & $\times$ & $\times$ & $\times$ & $\times$ & $\times$ & $\times$\\
\hline
\multirow{2}{*}{\tabincell{l}{Speech recog-\\nition}} & \tabincell{l}{Train} & \CheckmarkBold & \CheckmarkBold & \CheckmarkBold & $\times$ & $\times$ & $\times$ & \CheckmarkBold\\
\cline{2-9}
& \tabincell{l}{Infer} & \CheckmarkBold & \CheckmarkBold & \CheckmarkBold & $\times$ & $\times$ & $\times$ & \CheckmarkBold\\
\hline
\multirow{2}{*}{\tabincell{l}{Face \\embedding}} & \tabincell{l}{Train} & \CheckmarkBold & $\times$ & $\times$ & $\times$ & $\times$ & $\times$ & $\times$\\
\cline{2-9}
& \tabincell{l}{Infer} & \CheckmarkBold & $\times$ & $\times$ & $\times$ & $\times$ & $\times$ & $\times$\\
\hline
\multirow{2}{*}{\tabincell{l}{3D Face \\Recognition}} & \tabincell{l}{Train} & \CheckmarkBold & $\times$ & $\times$ & $\times$ & $\times$ & $\times$ & $\times$\\
\cline{2-9}
& \tabincell{l}{Infer} & \CheckmarkBold & $\times$ & $\times$ & $\times$ & $\times$ & $\times$& $\times$\\
\hline
\multirow{2}{*}{\tabincell{l}{Object \\detection}} & \tabincell{l}{Train} & \CheckmarkBold & \CheckmarkBold & $\times$ & $\times$ & $\times$ & $\times$ & \CheckmarkBold \\
\cline{2-9}
& \tabincell{l}{Infer} & \CheckmarkBold & \CheckmarkBold & $\times$ & $\times$ & $\times$ & $\times$ & $\times$\\
\hline
\multirow{2}{*}{\tabincell{l}{Recommendation}} & \tabincell{l}{Train} & \CheckmarkBold & \CheckmarkBold & $\times$ & $\times$ & $\times$ & $\times$ & \CheckmarkBold\\
\cline{2-9}
& \tabincell{l}{Infer} & \CheckmarkBold & $\times$ & $\times$ & $\times$ & $\times$ & $\times$& $\times$\\
\hline
\multirow{2}{*}{\tabincell{l}{Video \\prediction}} & \tabincell{l}{Train} & \CheckmarkBold & $\times$ & $\times$ &$\times$ & $\times$ & $\times$ & $\times$\\
\cline{2-9}
& \tabincell{l}{Infer} & \CheckmarkBold & $\times$ & $\times$ & $\times$ & $\times$ & $\times$ &$\times$\\
\hline
\multirow{2}{*}{\tabincell{l}{Image \\compression}} & \tabincell{l}{Train} & \CheckmarkBold & $\times$ & \CheckmarkBold & $\times$ & $\times$ & $\times$&$\times$ \\
\cline{2-9}
& \tabincell{l}{Infer} & \CheckmarkBold & $\times$ & \CheckmarkBold & $\times$ & $\times$ & $\times$ &$\times$\\
\hline
\multirow{2}{*}{\tabincell{l}{3D object re-\\construction}} & \tabincell{l}{Train} & \CheckmarkBold & $\times$ & $\times$ & $\times$ & $\times$ & $\times$ &$\times$\\
\cline{2-9}
& \tabincell{l}{Infer} & \CheckmarkBold & $\times$ & $\times$ & $\times$ & $\times$ & $\times$ &$\times$\\
\hline
\multirow{2}{*}{\tabincell{l}{Text sum-\\marization}} & \tabincell{l}{Train} & \CheckmarkBold & $\times$ & $\times$ & $\times$ & $\times$ & $\times$ &$\times$\\
\cline{2-9}
& \tabincell{l}{Infer} & \CheckmarkBold & $\times$ & $\times$ & $\times$ & $\times$ & $\times$ &$\times$\\
\hline
\multirow{2}{*}{\tabincell{l}{Spatial\\ transformer}} & \tabincell{l}{Train} & \CheckmarkBold & $\times$ & $\times$ & $\times$ & $\times$ & $\times$ &$\times$\\
\cline{2-9}
& \tabincell{l}{Infer} & \CheckmarkBold & $\times$ & $\times$ & $\times$ & $\times$ & $\times$ &$\times$\\
\hline
\multirow{2}{*}{\tabincell{l}{Learning to rank}} & \tabincell{l}{Train} & \CheckmarkBold & $\times$ & $\times$ & $\times$ & $\times$ & $\times$ &$\times$\\
\cline{2-9}
& \tabincell{l}{Infer} & \CheckmarkBold & $\times$ & $\times$ & $\times$ & $\times$ & $\times$& $\times$\\
\hline
\multirow{2}{*}{\tabincell{l}{Games}} & \tabincell{l}{Train} & $\times$ & \CheckmarkBold & \CheckmarkBold & $\times$ & $\times$ & $\times$ &\CheckmarkBold\\
\cline{2-9}
& \tabincell{l}{Infer} & $\times$ & $\times$ & \CheckmarkBold & $\times$ & $\times$ & $\times$ &$\times$\\
\hline
\multirow{2}{*}{\tabincell{l}{Memory network}} & \tabincell{l}{Train} & $\times$ & $\times$ & \CheckmarkBold & $\times$ & $\times$ & $\times$ & $\times$\\
\cline{2-9}
& \tabincell{l}{Infer} & $\times$ & $\times$ & \CheckmarkBold & $\times$ & $\times$ & $\times$ &$\times$\\
\hline
\multirow{2}{*}{\tabincell{l}{Question\\ answering}} & \tabincell{l}{Train} & $\times$ & $\times$ & $\times$ & $\times$ & $\times$ & \CheckmarkBold &$\times$\\
\cline{2-9}
& \tabincell{l}{Infer} & $\times$ & $\times$ & $\times$ & $\times$ & $\times$ & \CheckmarkBold &$\times$\\
\hline
\rowcolor{mygray} \multicolumn{9}{|l|}{Micro Benchmark}\\
\hline
%\multicolumn{2}{|c|}{Number} & 10 & N/A & N/A & 4 & 8 & N/A \\
%\hline
\multicolumn{2}{|c|}{Convolution} & \CheckmarkBold & $\times$ & $\times$ & \CheckmarkBold & \CheckmarkBold & $\times$ & $\times$\\
\hline
\multicolumn{2}{|c|}{Fully connected} & \CheckmarkBold & $\times$ & $\times$ & \CheckmarkBold & \CheckmarkBold & $\times$ &$\times$\\
\hline
\multicolumn{2}{|c|}{\tabincell{c}{Element-wise op}} & \CheckmarkBold & $\times$ & $\times$ & $\times$ & $\times$ & $\times$ &$\times$\\
\hline
\multicolumn{2}{|c|}{Activation} & \CheckmarkBold & $\times$ & $\times$ & $\times$ & \CheckmarkBold & $\times$ &$\times$\\
\hline
%\multicolumn{2}{|c|}{Sigmoid} & \CheckmarkBold & $\times$ & $\times$ & \CheckmarkBold & \CheckmarkBold & $\times$ \\
%\hline
%\multicolumn{2}{|c|}{Tanh} & \CheckmarkBold & $\times$ & $\times$ & \CheckmarkBold & \CheckmarkBold & $\times$ \\
%\hline
\multicolumn{2}{|c|}{Pooling} & \CheckmarkBold & $\times$ & $\times$ & $\times$ & \CheckmarkBold & $\times$ &$\times$\\
\hline
%\multicolumn{2}{|c|}{AvgPooling} & \CheckmarkBold & $\times$ & $\times$ & \CheckmarkBold & \CheckmarkBold & $\times$ \\
%\hline
\multicolumn{2}{|c|}{Normalization} & \CheckmarkBold & $\times$ & $\times$ & $\times$ & \CheckmarkBold & $\times$ &$\times$\\
\hline
%\multicolumn{2}{|c|}{BatchNorm} & \CheckmarkBold & $\times$ & $\times$ & \CheckmarkBold & \CheckmarkBold & $\times$ \\
%\hline
\multicolumn{2}{|c|}{Dropout} & \CheckmarkBold & $\times$ & $\times$ & $\times$ & \CheckmarkBold & $\times$ &$\times$\\
\hline
\multicolumn{2}{|c|}{Softmax} & \CheckmarkBold & $\times$ & $\times$ & $\times$ & \CheckmarkBold & $\times$ &$\times$\\
\hline
\multicolumn{2}{|c|}{AllReduce} & $\times$ & $\times$ & $\times$ & \CheckmarkBold & $\times$ & $\times$ &$\times$\\
\hline
\rowcolor{mygray} \multicolumn{9}{|l|}{Real-world Data sets and Software Stack}\\
\hline
\multicolumn{2}{|c|}{Text data} & 3 & 1 & 2 & N/A & N/A & 1 & 1\\
\hline
\multicolumn{2}{|c|}{Image data} & 8 & 2 & 2 & N/A & N/A & 2& 4 \\
\hline
\multicolumn{2}{|c|}{3D data} & 2 & 0 & 0 & N/A & N/A & 0 & 0\\
\hline
\multicolumn{2}{|c|}{Audio data} & 1 & 0 & 1 & N/A & N/A & 0 & 2\\
\hline
\multicolumn{2}{|c|}{Video data} & 1 & 0 & 1 & N/A & N/A & 0 & 0\\
\hline
\multicolumn{2}{|c|}{Software Stack} & 3 & 2 & 1 & 1 & 1 & 2 & 4\\
\hline

\end{tabular}
\end{table}

MLPerf~\cite{mlperf} is an ML benchmark suite targeting six AI problem domains, including image classification, object detection, translation, recommendation, speech recognition, and reinforcement learning. For several problem domains, it provides both light-weight and heavy-weight implementations.  Totally, it includes seven benchmarks for training and five benchmarks for inference. 
%This benchmark suite only covers five problem domains and provides component benchmarks with the lack of micro benchmarks and end-to-end application benchmarks.

Fathom~\cite{adolf2016fathom} provides eight deep learning component benchmarks implemented with TensorFlow, among which, three of them use different models to solve image classification problem. The Autoenc workload provides a variational autoencoder and can be used to reduce the dimension and compress images. This benchmark suite also lacks of the micro and application benchmarks. %and only implements the benchmarks using the TensorFlow stack.

DeepBench~\cite{deepbench} consists of four operations involved in training deep neural networks, including three basic operations and recurrent layer types. Although the recurrent layer is the combination of several basic operations like convolution, while it is still simpler comparing to our component benchmarks.
In total,  this benchmark suite only provides micro benchmarks, and lacks of the component and application benchmarks.

%BenchNN~\cite{chen2012benchnn} develops and evaluates software neural network implementations of 5 (out of 12) high-performance applications from the PARSEC Benchmark Suite. This benchmark only reimplement several PARSEC benchmarks using one software stack.

DNNMark~\cite{dong2017dnnmark} is a GPU benchmark suite that consists of a collection of deep neural network primitives. It provides eight micro benchmarks while lacking of component and application benchmarks.
%Tonic Suite~\cite{hauswald2015djinn} presents seven neural network workloads that use the DjiNN service.

DAWNBench~\cite{coleman2017dawnbench} is a benchmark and competition focusing on end-to-end performance, which means the training or inference time to achieve a state-of-the-art accuracy. It only focuses on two component benchmarks including image classification on CIFAR10 and ImageNet, and question answering on SQuAD, and lacks of the micro and application benchmarks.

TBD Suite~\cite{zhu2018tbd} is a benchmark suite for DNN training. It provides eight neural network models that covers six application domains. Analogously, it only contains component-level benchmarks, and lacks of an end-to-end application benchmark that can depict the entire execution paths of industry scale application.

Additionally, for machine learning and deep learning evaluation, MLModelScope~\cite{dakkak2019model} proposes a specification for repeatable model evaluation and a runtime to measure experiments.

In conclusion, the state-of-the-art and state-of-the-practise AI benchmarks only provide a few micro or component benchmarks, and none of them is
able to characterize an industry-scale Internet service.  AIBench provides a benchmark framework that all benchmarks collectively constitute an end-to-end application benchmark with an underlying industry scale Internet service model, while each individually forms a  micro or component benchmark that supports fine-grained benchmarking for AI systems, architectures, and algorithms. 
Meanwhile, the HPC AI benchmarks~\cite{jiang2018hpc}, IoT AI benchmarks~\cite{luo2018iot}, Edge AI benchmarks~\cite{hao2018edge}, the previous version of AIBench for datacenter~\cite{aibench}, and big data benchmarks~\cite{gao2018bigdatabench,wang2014bigdatabench,jia2013characterizing} are also released on the BenchCouncil web site.

\section{AIBench Framework}\label{framework}

In this section, we introduce the AIBench framework from the perspectives of framework architecture, metrics, prominent AI problem domains we identified, implemented micro and component benchmarks, and framework scalability on large-scale clusters.

\subsection{Framework Architecture}

\begin{figure}[tb]
\centering
\includegraphics*[scale=1]{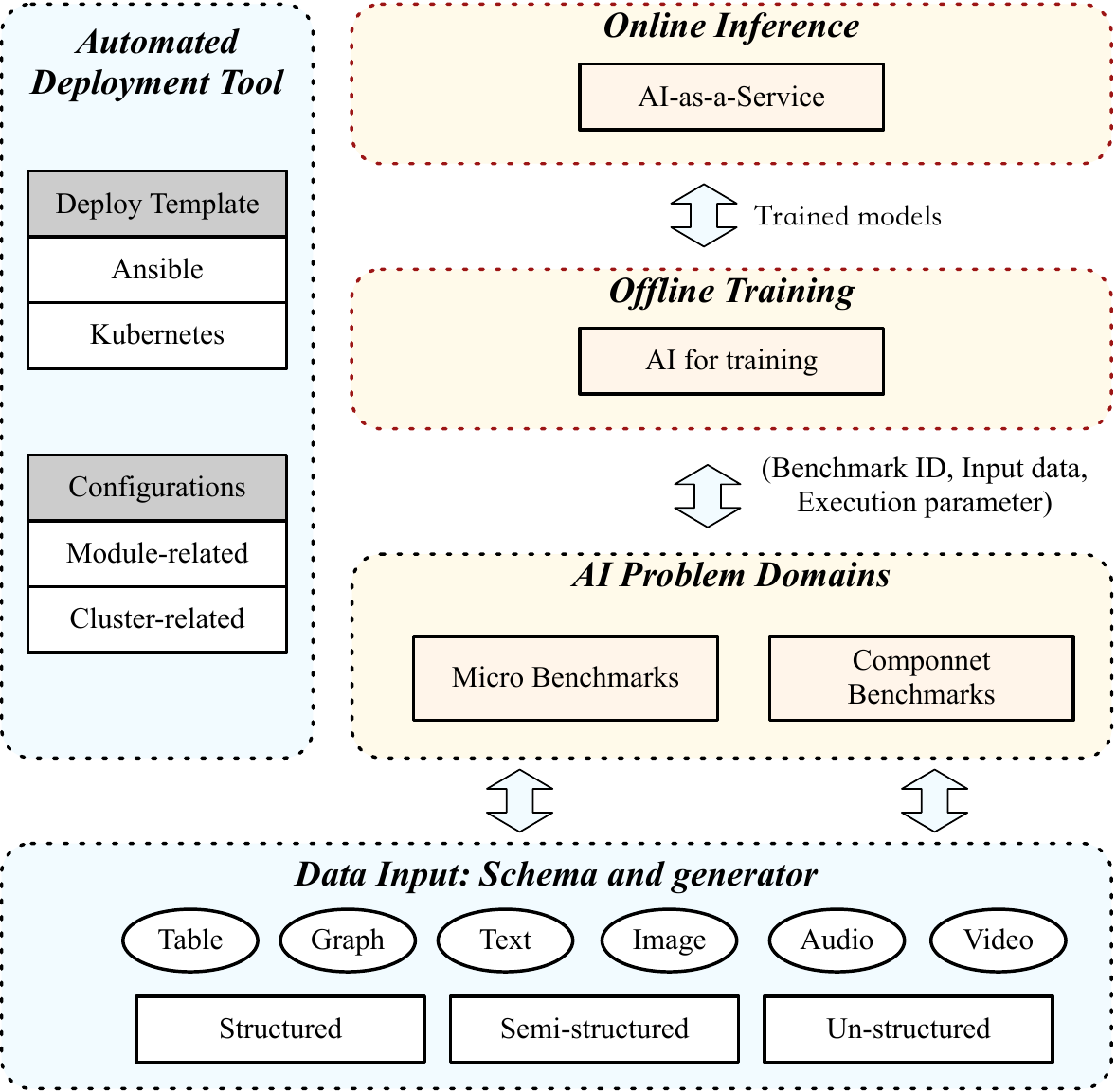}
\caption{AIBench Framework.} %\vspace{1pt}
\label{arch-design}
\end{figure}

The AIBench framework provides an universal AI benchmark framework that is flexible and configurable, which is shown in Fig.~\ref{arch-design}. 
It provides loosely coupled modules that can be easily configured and extended to compose an end-to-end application, including the data input, AI problem domain, online inference, offline training, and deployment tool modules. 

% to constitute the entire lifecycle of an internet service application

The data input module is responsible to feed data into the other modules. It collects representative real-world data sets from  not only  the authoritative public websites but also our industry partners after anonymization.  The data schema is designed to maintain the real-world data characteristics, so as to alleviate the confidential issue. Based on the data schema, a series of data generators are further provided to support large-scale data generation, like the user or product information. To  cover a wide spectrum of data characteristics, we take diverse data types, e.g., structured, semi-structured, un-structured, and different data sources, e.g., table, graph, text, image, audio, video, into account. Our framework integrates various open-source data storage systems, and supports large-scale data generation and deployment~\cite{ming2014bdgs}.

To achieve diversity and representativeness of our framework, we first identify prominent  AI problem domains that play important roles in most important Internet services domains. And then we provide the concrete implementation of the AI algorithms targeting those AI problem domains as component benchmarks. 
Also, we profile the most intensive units of computation across those component benchmarks, and implement them  as a set of micro benchmarks. Both micro and component benchmarks are implemented with the concern for composability, each of which can run collectively and individually.

The offline training and online inference modules are provided to construct an end-to-end application benchmark. First, the offline training module chooses one or more component benchmarks from the AI problem domain module, through specifying the required benchmark ID, input data, and execution parameters like batch size. Then the offline training module trains a model and provides the trained model to the online inference module. The online inference module loads the trained model onto the serving system, i.e., TensorFlow serving. Collaborating with the other non AI-related modules in the critical paths, an end-to-end application benchmark is built.

To be easily deployed on a large-scale cluster, the framework provides deployment tools that contain two automated deployment templates using Ansible and Kubernetes, respectively. Among them, the Ansible templates support scalable deployment on physical machines or virtual machines, while the kubernetes templates are used to deploy on container clusters. A configuration file needs to be specified for installation and deployment, including module parameters like the chosen benchmark ID, input data, and the cluster parameters like nodes, memory, and network information.

\subsection{The Prominent AI Problem Domains}\label{identify}

%We first introduction the problem domains identified from industries, including internet service, intelligence medicine, recognition science, etc. Then we illustrate a series of micro and application benchmarks that serve as a service or training model in AIBench framework. 

To cover a wide spectrum of prominent AI problem domains among Internet services, we thoroughly analyze the core scenarios among three primary Internet services, including search engine, social network, and e-commerce, as shown in Table~\ref{problemdomain}. In total, we identify  sixteen representative AI problem domains as follows.

\begin{table*}[htbp]
%\scriptsize
\caption{Prominent AI Problem Domains in Internet Service.}
\renewcommand\arraystretch{1.05}
\scriptsize
\label{problemdomain}
\center %p{0.455in}|
\begin{tabular}{|p{0.85in}|p{2.4in}|p{2.6in}|}
\hline
\textbf{Internet Service} & \textbf{Core Scenario} & \textbf{Involved AI Problem Domain} \\
\hline
\multirow{9}{*}{Search Engine} & \multirow{2}{*}{Content-based image retrieval (e.g., face, scene)} & Object detection; Classification; Spatial transformer; Face embedding; 3D face recognition \\
\cline{2-3}
& Advertising and recommendation & Recommendation \\
\cline{2-3}
& \multirow{2}{*}{Maps search and translation} & 3D object reconstruction; Text-to-Text translation; Speech recognition \\
\cline{2-3}
& Data annotation and caption (e.g., text, image) & Text summarization; Image-to-Text \\
\cline{2-3}
& Search result ranking & Learning to rank \\
\cline{2-3}
& Image resolution enhancement & Image generation; Image-to-Image \\
\cline{2-3}
& Data storage space and transfer optimization & Image compression; Video prediction \\
\hline

\multirow{9}{*}{Social Network} & Friend or community recommendation & Recommendation; Face embedding; 3D face recognition; \\
\cline{2-3}
& Vertical search (e.g., image, people) & Classification; Spatial transformer;  Object detection;  \\
\cline{2-3}
& Language translation & Text-to-Text translation \\
\cline{2-3}
& Automated data annotation and caption & Text summarization; Image-to-Text; Speech recognition \\
\cline{2-3}
& Anomaly detection (e.g., spam image detection) & Classification \\
\cline{2-3}
& Image resolution enhancement & Image generation; Image-to-Image \\
\cline{2-3}
& Photogrammetry (3D scanning) & 3D object reconstruction \\
\cline{2-3}
& Data storage space and transfer optimization & Image compression; Video prediction \\
\cline{2-3}
& News feed ranking & Learning to rank \\
\hline

\multirow{9}{*}{E-commerce} & Product searching & Classification; Spatial transformer;  Object detection  \\
\cline{2-3}
& Product recommendation and advertising & Recommendation \\
\cline{2-3}
& Language and dialogue translation & Text-to-Text translation; Speech recognition \\
\cline{2-3}
& Automated data annotation and caption & Text summarization; Image-to-Text  \\
\cline{2-3}
& \multirow{2}{*}{Virtual reality (e.g., virtual fitting)} & 3D object reconstruction; Image generation; Image-to-Image \\
\cline{2-3}
& Data storage space and transfer optimization & Image compression; Video prediction \\
\cline{2-3}
& Product ranking & Learning to rank \\
\cline{2-3} 
& Facial authentication and payment & Face embedding; 3D face recognition; \\
\hline

\end{tabular}
\end{table*}

\textbf{Classification.} This problem domain is to extract different thematic classes within the input data like an image or a text file, which is a supervised learning problem to define a set of target classes and train a model to recognize. It is a typical task in Internet services or other application domains, and widely used in multiple scenarios, like category prediction and spam detection.

\textbf{Image Generation.} This problem domain aims to provide an unsupervised learning problem to mimic the distribution of data and generate images. The typical scenario of this task includes image resolution enhancement, which can be used to generate high-resolution image.

\textbf{Text-to-Text Translation.} This problem domain need to translate text from one language to another, which is the most important field of computational linguistics. It can be used to translate the search query intelligently and translate dialogue.

\textbf{Image-to-Text.} This problem domain is to generate the description of an image automatically. It can be used to generate image caption and recognize optical character within an image.

\textbf{Image-to-Image.} This problem domain is to convert an image from one representation of an image to another representation. It can be used to synthesize the images with different facial ages and simulate virtual makeup. Face aging can help search the facial images ranging different age stages.

\textbf{Speech recognition.} This problem domain is to recognize and translate the spoken language to text. This task is beneficial for voice search and voice dialogue translation.

\textbf{Face embedding.} This problem domain is to transform a facial image to a vector in embedding space. The typical scenarios of this task are facial similarity analysis and face recognition.

\textbf{3D face recognition.} This problem domain is to recognize the 3D facial information from multiple images from different angles. This task mainly focuses on three-dimensional images and is beneficial to the facial similarity and facial authentication scenario.

\textbf{Object detection.} This problem domain is to detect the objects within an image. The typical scenarios are vertical search like contented-based image retrieval and video object detection.

\textbf{Recommendation.} This problem domain is to provide recommendations. This task is widely used for advertise recommendation, community recommendation, or product recommendation.

\textbf{Video prediction.} This problem domain is to predict the future video frames through predicting previous frames transformation. The typical scenarios are video compression and video encoding, for efficient video storage and transmission.

\textbf{Image compression.} This problem domain is to compress the images and reduce the redundancy~\cite{toderici2017full}. The task is important for Internet service in terms of data storage overhead and data transmission efficiency.

\textbf{3D object reconstruction.} This problem domain is to predict and reconstruct 3D objects~\cite{yan2016perspective}. The typical scenarios are maps search, light field rendering and virtual reality.

\textbf{Text summarization.} This problem domain is to generate the text  summary, which is important for search results preview, headline generation, and keyword discovery.

\textbf{Spatial transformer.} This problem domain is to perform spatial transformations~\cite{jaderberg2015spatial}. An typical scenario of this task is space invariance image retrieval, so that the image can be retrieved even if the image is extremely stretched.

\textbf{Learning to rank.} This problem domain is to learn the attributes of searched content and rank the scores for the results, which is the key for searching service.

%To cover a full spectrum of data characteristics, 

\subsection{Micro and Component Benchmarks}

Targeting  the major AI problem domains abstracted in Section~\ref{identify}, we provide the concrete implementation of the AI algorithms. Individually, these algorithm implementations form a series of micro and component benchmarks for fine-grained evaluation. Table~\ref{AIBench_component} and Table~\ref{AIBench_micro} list the component and micro benchmarks in AIBench. In total, AIBench includes sixteen component benchmarks for AI problem domains and twelve micro benchmarks that extract unit of computation from the typical AI algorithms~\cite{gao2018motif}.

\begin{table*}[htbp]
\scriptsize
\caption{Component Benchmarks in AIBench.}
\renewcommand\arraystretch{1.4}
\label{AIBench_component}
\center %p{0.455in}|
\begin{tabular}{|p{0.65in}|p{1.36in}|p{2in}|p{1.6in}|}
\hline
\textbf{No.} & \textbf{Component Benchmark} & \textbf{Algorithm} & \textbf{Data Set}\\
\hline
DC-AI-C1 & Image classification & ResNet50~\cite{he2016deep} &  ImageNet \\
\hline
DC-AI-C2 & Image generation & WassersteinGAN~\cite{arjovsky2017wasserstein} &  LSUN \\
\hline
DC-AI-C3 & Text-to-Text translation & Transformer~\cite{vaswani2017attention} &  WMT English-German \\
\hline
DC-AI-C4 & Image-to-Text & Neural Image Caption Model~\cite{vinyals2017show} &  Microsoft COCO \\
\hline
DC-AI-C5 & Image-to-Image & CycleGAN~\cite{zhu2017unpaired} &  Cityscapes \\
\hline
DC-AI-C6 & Speech recognition & DeepSpeech2~\cite{amodei2016deep} &  Librispeech \\
\hline
DC-AI-C7 & Face embedding & Facenet~\cite{schroff2015facenet} &  LFW, VGGFace2 \\
\hline
DC-AI-C8 & 3D Face Recognition & 3D face models~\cite{vieriu2015facial} &  77,715 samples from 253 face IDs\\
\hline
DC-AI-C9 & Object detection & Faster R-CNN~\cite{ren2015faster} &  Microsoft COCO \\
\hline
DC-AI-C10 & Recommendation & Neural collaborative filtering~\cite{he2017neural} &  MovieLens \\
\hline
DC-AI-C11 & Video prediction & Motion-Focused predictive models~\cite{finn2016unsupervised} &  Robot pushing data set \\
\hline
DC-AI-C12 & Image compression & Recurrent neural network~\cite{toderici2017full} &  ImageNet \\
\hline
DC-AI-C13 & 3D object reconstruction & Convolutional encoder-decoder network~\cite{yan2016perspective} &  ShapeNet Data set\\
\hline
DC-AI-C14 & Text summarization & Sequence-to-sequence model~\cite{nallapati2016abstractive} &  Gigaword data set \\
\hline
DC-AI-C15 & Spatial transformer & Spatial transformer networks~\cite{jaderberg2015spatial} &  MNIST \\
\hline
DC-AI-C16 & Learning to rank & Ranking distillation~\cite{tang2018ranking} &  Gowalla \\
\hline

\end{tabular}
\end{table*}

\begin{table}[htbp]
%\scriptsize
\caption{Micro Benchmarks in AIBench.}
\renewcommand\arraystretch{1.2}
\scriptsize
\label{AIBench_micro}
\center %p{0.455in}|
\begin{tabular}{|p{1in}|p{2in}|p{2in}|}
\hline
\textbf{No.} & \textbf{Micro Benchmark} & \textbf{Data Set} \\
\hline
DC-AI-M1 & Convolution  & ImageNet, Cifar \\
\hline
DC-AI-M2 & Fully Connected & ImageNet, Cifar \\
\hline
DC-AI-M3 & Relu  & ImageNet, Cifar \\
\hline
DC-AI-M4 & Sigmoid  & ImageNet, Cifar \\
\hline
DC-AI-M5 & Tanh  & ImageNet, Cifar \\
%\hline
%Tanh & Matrix & SN, EC, MP, BI & AI & Cifar, ImageNet & TensorFlow, Caffe, PyTorch\\
\hline
DC-AI-M6 & MaxPooling   & ImageNet, Cifar \\
\hline
DC-AI-M7 & AvgPooling & ImageNet, Cifar \\
\hline
DC-AI-M8 & CosineNorm~\cite{luo2017cosine}  &  ImageNet, Cifar\\
\hline
DC-AI-M9 & BatchNorm~\cite{ioffe2015batch}   &  ImageNet, Cifar \\
\hline
DC-AI-M10 & Dropout~\cite{srivastava2014dropout}  &  ImageNet, Cifar \\
%\cline{1-1}\cline{3-7}
%Eltwise & & Matrix & SN, MP, BI & AI & Cifar, ImageNet & TensorFlow, Caffe\\
\hline
DC-AI-M11 & Element-wise multiply  &  ImageNet, Cifar \\
\hline
DC-AI-M12 & Softmax  &  ImageNet, Cifar \\
\hline

\end{tabular}
\end{table}

%\subsection{Scalability on Large-scale Clusters}

\subsection{Data Model}

To cover a diversity of data sets from various applications, we collect 15 representative data sets, including ImageNet~\cite{deng2009imagenet}, Cifar~\cite{krizhevsky2014cifar}, LSUN~\cite{yu2015lsun}, WMT English-German~\cite{wmt}, Cityscapes~\cite{cordts2016cityscapes}, LibriSpeech~\cite{panayotov2015librispeech}, Microsoft COCO data set~\cite{lin2014microsoft}, LFW~\cite{huang2008labeled}, VGGFace2~\cite{cao2018vggface2}, Robot pushing data set~\cite{finn2016unsupervised},  MovieLens data set~\cite{harper2016movielens}, ShapeNet data set~\cite{chang2015shapenet}, Gigaword data set~\cite{rush2017neural}, MNIST data set~\cite{lecun2010mnist}, Gowalla data set~\cite{cho2011friendship} and the 3D face recognition data set from our industry partner.

\subsection{Metrics}

AIBench focuses on a series of metrics covering accuracy, performance, and energy consumption, which are major industry concerns.

The metrics for online inference contains query response latency, tail latency, and throughput from performance aspect, inference accuracy, and inference energy consumption.

The metrics for offline training contains the samples processed per second, the wall clock time to train the specific epochs, the wall clock time to train a model achieving a target accuracy~\cite{coleman2017dawnbench}, and the energy consumption to train a model achieving a target accuracy~\cite{coleman2017dawnbench}.

\section{The Design and Implementation of Application Benchmark}\label{methodology}

On the basis of the AIBench framework illustrated in Section~\ref{framework}, we implement the first end-to-end AI application benchmark, modelling the complete use-cases of a realistic E-commerce search intelligence.

\subsection{The Design and Implementation}

\begin{figure}[tb]
\centering
\includegraphics*[scale=0.68]{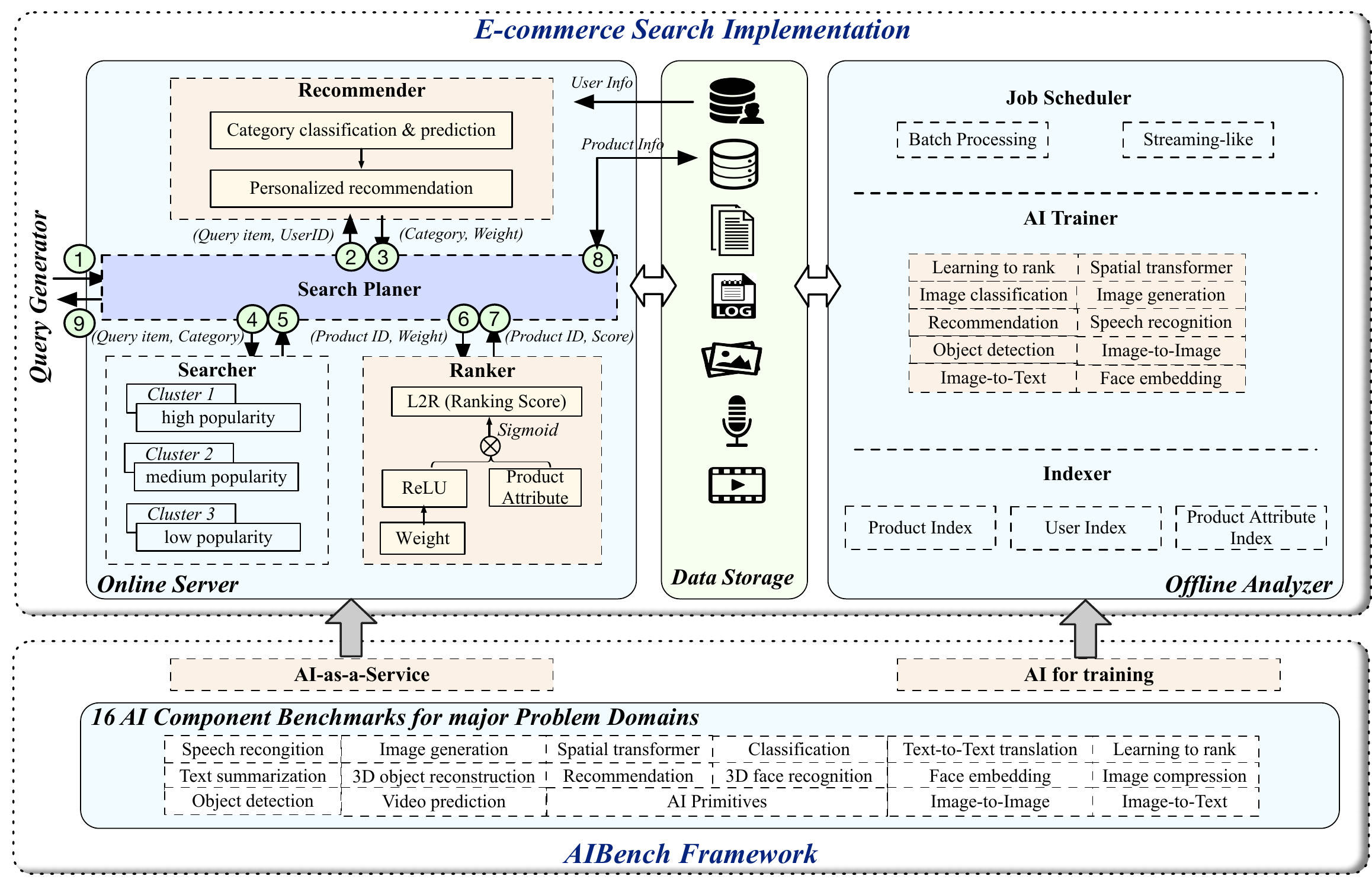}
\caption{AIBench Implementation.} %\vspace{1pt}
\label{ecommerce-arch}
\end{figure}

The end-to-end application benchmark consists of four modules: online server, offline analyzer, query generator, and data storage, as shown in Fig.~\ref{ecommerce-arch}. Among them, online server receives the query requests and performs personalized searching and recommendation, integrated with AI inference. %, which is a online searching process integrated with AI inference. 

Offline analyzer chooses the  appropriate AI  algorithm implementations and performs a training stage to generate a learning model. Also, the offline analyzer is responsible to build data indexes to accelerate data access.

Query generator is to simulate concurrent users and send query requests to online server based on a specific configuration. The configuration designates parameters like concurrency, query arriving rate,  distribution, and user thinking time, to simulate different query characteristics and satisfy multiple generation strategies. We implement our query generator based on JMeter~\cite{jmeter2017apache}.

Data storage module stores all kinds of data, including the user database that saves all the attributes of user information, the product database that holds all the attributes of product information, logs that record the complete query histories, text data that contains the product description text or the user comments, image and video data that depict the appearance and usage of product vividly, and audio data that stores the voice search data and voice chat data. Overall, the data storage covers various data types including structured, un-structured, and semi-structured data, and diverse data sources, including table, text, image, audio and video.

To support scalable deployment on the clusters with different scales, each module is able to deploy on multiple nodes. Also, a series of data generators are provided to generate the e-commerce data with different scales, through setting several parameters, e.g., the number of products and product attribute fields, the number of users and user attribute fields.

\subsubsection{Online Server}

Online server provides personalized searching and recommendations combining traditional machine learning and deep learning technologies. Online server consists of four submodules including search planer, recommender, searcher, and ranker.

\textbf{Search Planer} is the entrance of online server.  It is responsible for receiving the query requests from query generator, and sending the request to the other online components and receiving the return results. We use the Spring Boot framework~\cite{webb2013spring} to implement the search planer.

\textbf{Recommender} is to analyze the query item and provides personalized recommendation, according to the user information obtained from the user database. It first conducts query spelling correction and query rewrite, then it predicts the belonged category of the query item based on a classification model---FastText~\cite{joulin2016fasttext}. Using a deep neural network proposed by Alibaba ~\cite{ni2018perceive}, the query planer then conducts an inference process and uses the offline trained model to provide personalized recommendation. It returns two vectors---the probability vector of the predicted categories and the user preference score vector of product attribute, such as the user preference for brand, color, etc. 
We use the Flask Web framework~\cite{grinberg2018flask} and Nginx~\cite{reese2008nginx} to build our recommender for category prediction, and adopt TensorFlow serving~\cite{olston2017tensorflow} to implement online recommendation.

To guarantee scalability and service efficiency, Searcher follows an industry scale architecture. \textbf{Searcher} is deployed on several different, i.e., three  clusters, that hold the inverted indexes of product information in memory to guarantee high concurrency and low latency. In view of the click-through rate and purchase rate, the products belong to three categories according to the popularity---high, medium, and low, occupying the proportion of 15\%, 50\%, and 50\%, respectively. Note that the high popularity category is the top 15\% popular products chosen from the medium popularity category.  The indexes of products with different popularities are stored into different clusters.  
Given a searching request, the searcher searches these three clusters one by one, until reaching a specific amount. Generally, the cluster that holds low popularity products is rarely searched in a realistic scenario.
So for each category, searcher adopts different deployment strategies. The cluster for high popularity contains more nodes and more backups to guarantee the searching efficiency. While the cluster for low popularity deploys the least number of nodes and backups. We use the Elasticsearch~\cite{gormley2015elasticsearch} to set up and manage the three clusters of searcher.
%Considering the popularity of goods, there are three individual clusters for goods searching with different popularities.

\textbf{Ranker} uses the weight returned by \emph{Recommender} as initial weight, and ranks the scores of products through a personalized L2R neural network~\cite{ni2018perceive}. The ranker also uses the  Elasticsearch~\cite{gormley2015elasticsearch} to implement product ranking. %\emph{Goods Database} stores the whole information of all goods. 

The online serving process is as follows.

(1) \emph{Query Generator} simulates concurrent users and sends query requests to \emph{Search Planer};

(2) \emph{Search Planer} receives the query request and sends the query item to \emph{Recommender};

(3) \emph{Recommender} analyzes the query and returns the category prediction results and personalized attribute weights to \emph{Search Planer};

(4) \emph{Search Planer} sends the initial query item and predicted category results to \emph{Searcher};

(5) \emph{Searcher} searches the inverted indexes and returns the product ID to \emph{Search Planer};

(6) \emph{Search Planer} sends the product ID and personalized attribute weights to \emph{Ranker};

(7) \emph{Ranker} ranks the product according to the initial weights and returns the ranking scores to \emph{Search Planer};

(8) \emph{Search Planer} sends a product database access request according to the product ID and obtains the product information;

(9) \emph{Search Planer} returns the searched product information to \emph{Query Generator}.

\subsubsection{Offline Analyzer}

Offline analyzer is responsible for training models and building indexes to improve the online serving performance. It consists of three parts---AI trainer, job scheduler, and indexer.

AI trainer is to train models using related data stored in database. To dig the features within product data, e.g., text, image, audio, video, and power the efficiency of online server, the AI trainer chooses ten AI algorithms (component benchmarks) from the AIBench framework currently, including classification  for category prediction, recommendation  for personalized recommendation, learning to ranking  for result scoring and ranking, image-to-text  for image caption, image-to-image and image generation for image resolution enhancement, face embedding  for face detection within an image, spatial transformer  for image rotating and resizing, object detection  for detecting video data, speech recognition for audio data recognition. %  compression model for image compression, , and text summatization for product keyword generation.
 
Job schedule provides two kinds of training mechanisms: batch processing and streaming-like processing. In a realistic scenario, some models need to be updated frequently. For example, when we search an item and click one product showed in the first page, the application will immediately train a new model based on the product that we just clicked, and make new recommendations shown in the second page. Our benchmark implementation considers this situation, and adopts a streaming-like approach to update the model every several seconds. For batch processing, the trainer will update the models every several hours.
% while for streaming, the trainer will update the models every several seconds. 
The indexer is to build indexes for product information. In total, the indexer provides three kinds of indexes, including the inverted indexes with a few fields of products for searching, forward indexes with a few fields for ranking, and forward indexes with a majority of fields for summary generation.

\subsection{Extensibility for Other Industry Applications}

%The micro and component benchmarks provided by AIBench are not only suit for the construction of internet service applications, but also features prominently in many other industry applications. Moreover, the modules of AIBench are loosely coupled so that they can be extended to be an AI-related module in another critical path.

Taking medicine AI scenario as an example---which is one of the most representative scenario for AI, we illustrate how we use the AIBench framework to construct an end-to-end benchmark for clinical diagnosis application. 
The AI-related critical path of clinical diagnosis contains the following steps. 1) train a series of diagnosis models according to history data offline, e.g., detection model, classification model, recommendation model, etc; 2) detect the abnormal information within patient's physical examination data, such as the tumour detection of a CT image; 3) classify and predict the potential disease which is an online inference stage; 4) recommend an optimal treatment plan. 

To construct an end-to-end clinical diagnosis application benchmark, the AIBench framework is flexible to provide an AI-related offline module and online module. Within the offline module, component benchmarks of object detection, classification, and recommendation are chosen for training models. Within the online module, these models are loaded for online inference as a service.

In conclusion, the AIBench framework is extensible for building other industry applications.

\section{Experiment Setup}

In this section, we describe the experimental setting and methodology. %environment for our evaluations using AIBench, and the method that we obtain the performance data.

%run a series of characterization experiments using AIBench to obtain insights for architectural studies and explore the impact of different technologies on datacenter computing, like virtualization technology. In this section, we present our experiment configurations and methodology on obtaining performance data.

\subsection{Node Configurations}

We deploy a 16-node CPU and GPU cluster. 
For the CPU cluster, all the nodes are connected with a 1 Gb ethernet network. Each node is equipped with two Xeon E5645 processors and 32 GB memory. Each processor contains six physical out-of-order cores. The OS version of each node is Linux CentOS 6.9 with the Linux kernel version 3.11.10. The software versions are JDK 1.8.0, Python 3.6.8, and GCC 5.4, respectively. 
The GPU node is equipped with four Nvidia Titan XP GPUs. Every Titan XP owns 3840 Nvidia Cuda cores and 12 GB memory.
The detailed hardware configuration of each node is listed in Table~\ref{hwconfigeration}.

\begin{table}
\caption{Hardware Configuration Details.}\label{hwconfigeration}
\renewcommand\arraystretch{1.1}
\center
\scriptsize
\begin{tabular}{|p{1in}|p{1in}|p{1in}|p{1in}|}
\hline \rowcolor{mygray} \multicolumn{4}{|l|}{CPU Configurations}\\
\hline \multicolumn{2}{|c|}{CPU Type} & \multicolumn{2}{c|}{Intel CPU Core} \\
\hline \multicolumn{2}{|c|}{Intel \textregistered Xeon E5645}  &\multicolumn{2}{c|}{6 cores@2.40G} \\
\hline L1 DCache &L1 ICache &L2 Cache &L3 Cache \\
\hline 6 $\times$ 32 KB& 6 $\times$ 32 KB&6 $\times$ 256 KB& 12MB \\
%\hline \multicolumn{2}{|c|}{Memory} & \multicolumn{2}{c|}{32GB,DDR3}  \\
%\hline \multicolumn{2}{|c|}{Disk} & \multicolumn{2}{c|}{SATA@7200RPM}\\
%\hline \multicolumn{2}{|c|}{Ethernet} & \multicolumn{2}{c|}{1Gb}\\
\hline \multicolumn{2}{|c|}{Memory} & \multicolumn{2}{c|}{32GB, DDR3}\\
\hline \multicolumn{2}{|c|}{Ethernet} & \multicolumn{2}{c|}{1Gb}\\
\hline \multicolumn{2}{|c|}{Hyper-Threading} & \multicolumn{2}{c|}{Disabled}\\
%\hline
%\hline \rowcolor{mygray} \multicolumn{4}{|l|}{Software Configurations}\\
%\hline \tabincell{l}{Operating\\System} & \tabincell{l}{Linux%\\Kernel} & \tabincell{l}{JDK\\Version} & \tabincell{l}{Hadoop\\Version} \\
%\hline CentOS 6.4 & 3.11.10 & 1.7.0 & 2.7.1\\
\hline \rowcolor{mygray} \multicolumn{4}{|l|}{GPU Configurations}\\
\hline \multicolumn{2}{|c|}{GPU Type} & \multicolumn{2}{c|}{Nvidia Titan XP} \\
\hline \multicolumn{2}{|c|}{Nvidia Cuda Cores} & \multicolumn{2}{c|}{3840 cores} \\
\hline \multicolumn{2}{|c|}{GPU Memory} & \multicolumn{2}{c|}{12GB, GDDR5X} \\
\hline

\end{tabular}
\end{table}

\subsection{The Benchmark Deployment}

We deploy the end-to-end AI application benchmark, introduced in Section~\ref{methodology} on the above mentioned clusters, for online server and offline trainer.

\textbf{Online Server Setting.}
Online server is deployed on the 16-node CPU cluster, containing one query generator node, one search planer node, two recommender nodes, nine searcher nodes, one ranker node, and two nodes for data storage. The detailed module setup information and involved software information are listed in Table~\ref{server-setup}. 

\begin{table}
\caption{Online Server Settings.}\label{server-setup}
\renewcommand\arraystretch{1.1}
\center
\scriptsize
\begin{tabular}{|p{1in}|p{1.5in}|p{0.8in}|p{1.5in}|}
\hline \multicolumn{2}{|c|}{Module} &  Nodes & Software \\
\hline \multicolumn{2}{|c|}{Query Generator} & 1 & Jmeter 5.1.1  \\
\hline \multicolumn{2}{|c|}{Search Planer} & 1 & SpringBoot 2.1.3  \\
\hline \multirow{2}{*}{Recommender} & Category Predictor & 1 & Flask Web 1.1.1, Nginx 1.10.3   \\
\cline{2-4} & TensorFlow Serving & 1 & TensorFlow Serving 1.14.0 \\
\hline \multirow{3}{*}{Searcher} & Cluster (high popularity) & 4 & \multirow{3}{*}{Elasticsearch 6.5.2} \\
\cline{2-3} & Cluster (medium popularity) & 3 &  \\
\cline{2-3} & Cluster (low popularity) & 2 &  \\
\hline \multicolumn{2}{|c|}{Ranker} & 1 & Elasticsearch 6.5.2 \\
\hline \multirow{2}{*}{Data Storage} & User database & 1 & Neo4j 3.5.8  \\
\cline{2-4}  & Product database & 1 & Elasticsearch 6.5.2  \\
\hline

\end{tabular}
\end{table}

\textbf{Offline Trainer Settings.}
Offline trainer is deployed on the GPUs. The CUDA and Nvidia driver versions are 10.0 and 410.78, respectively. We evaluate the PyTorch implementations with the version of 1.1.0. The data sets for each benchmark are ImageNet for image classification (137 GB), LSUN for image generation (42.8 GB), VGGFace2 for face embedding (36 GB), Microsoft COCO (13 GB) for Image-to-Text and object detection, MNIST (9.5 MB) for spatial transformer, Cityscapes (267 MB) for Image-to-Image, MovieLens (190 MB)  for recommendation, Librispeech (59.3 GB) for speech recognition, and Gowalla (107 MB) for learning to rank, respectively.

\subsection{Performance Data Collection}

We use network time protocol (NTP)~\cite{mills1985network} to perform clock synchronization in all cluster nodes and obtain the latency and tail latency metrics of online server.
We use a profiling tool---Perf~\cite{de2010new} to collect the CPU micro-architectural data through hardware performance monitoring counters (PMCs). 
For GPU profiling, we use Nvidia profiling toolkit---nvprof~\cite{nvprof} to track the running performance of GPU.
To obtain more accurate metric numbers, we run each benchmark three times and report the average value.

\section{Evaluation}
In this section, we evaluate the end-to-end AI application benchmarks introduced in Section~\ref{methodology}, including online server and ten AI component benchmarks included in offline analyzer.

\subsection{Evaluation of Online Server}

We evaluate the online server performance on the 16-node CPU cluster. The product database contains a hundred thousand products with 32 attribute fields.
The query generator simulates 1000 users with 30-second warm up time. The users send query requests continuously every think time interval, which follows a poisson distribution. In total, we collect the performance numbers when 20,000 query requests finish.

%\subsubsection{Service Latency}

\begin{figure}[tb]
\centering
\includegraphics*[scale=0.53]{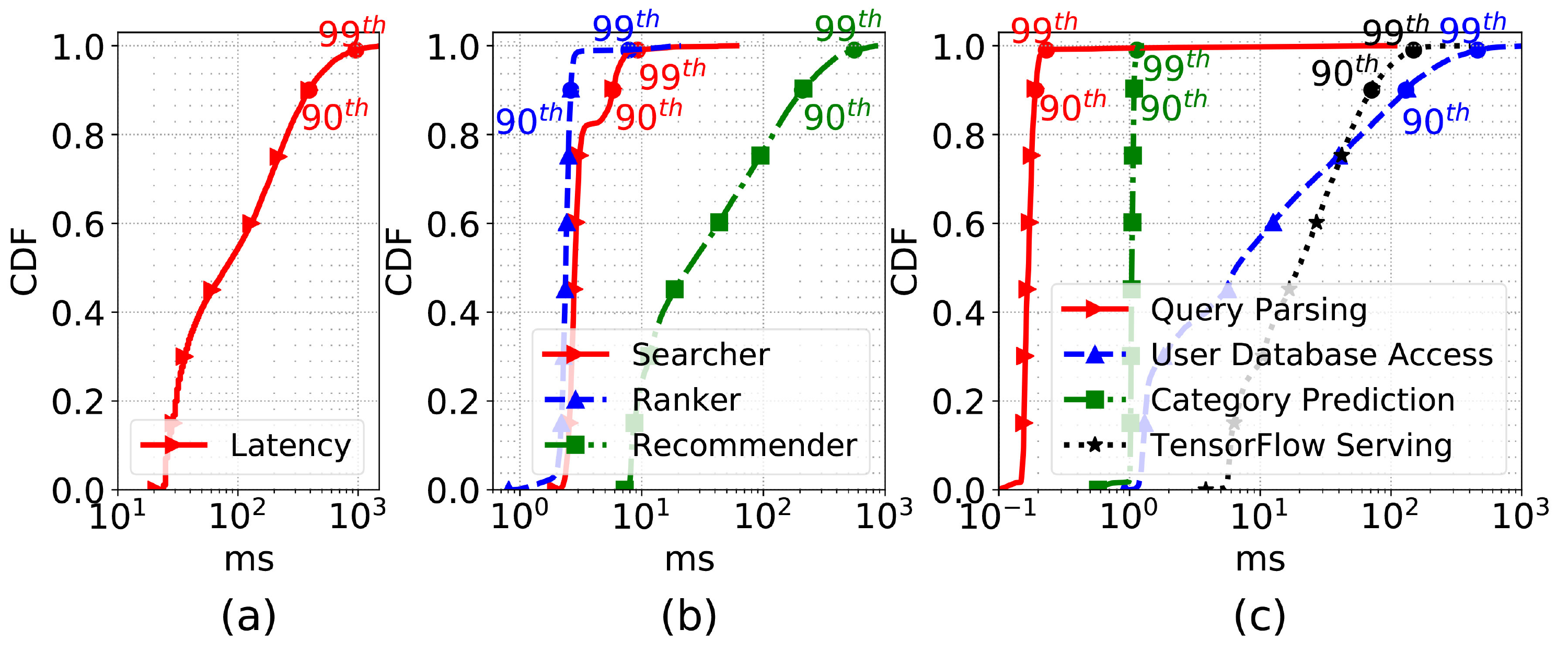}
\caption{Latency of Online Server.} %\vspace{1pt}
\label{latency}
\end{figure}

Latency is an important metric to evaluate the service quality.
Fig.~\ref{latency} shows the latencies of online server.  We find that the overall latencies of the entire execution paths of the current baseline implementation are 161.13, 392, and 956 milliseconds for the average, 90th percentile and 99th percentile latencies respectively, as listed in Fig.~\ref{latency}(a)~\footnote{With respect to the real numbers in our industry partner, the number is quite high. They have taken many measures to decrease the overall latency. However, this baseline implementation indeed confirm the importance of the AI components in the critical paths as the data access and communication overhead can be further decreased.}. 
We further deeply analyze the latency of each module, including recommender, searcher, and ranker, as shown in Fig.~\ref{latency}(b). The latency of search planer is negligible, so we do not report it in Fig.~\ref{latency}(b). We find that the recommender occupies the largest latency: 75.7 milliseconds, 209.4 milliseconds, and 557.7 milliseconds for the average, 90th percentile, 99th percentile latencies, respectively. In comparison, the latencies of searcher and ranker are both within 4 milliseconds. Although recommender and ranker both contain AI-related components, they incur different latencies. The average communication latency between the modules is also high, nearly equal to that of recommender.

Furthermore,  Fig.~\ref{latency}(c) presents the latency breakdown of recommender in terms of query parsing, user database access, category prediction, and TensorFlow serving.
We find that database access  and TensorFlow serving latencies  are the top two factors that impact the service performance. The user information is represented as graph data that contains entities and relationships. The sophisticated data structure and frequent garbage collection may influence the data access speed largely. While TensorFlow serving needs to run a forward inference using the recommendation model, and thus incurs larger latency. In order to measure the impact of the AI component on service performance and find the bottlenecks, we make a discussion from the following aspects.

\textbf{The weights of AI-related components on service performance.} AI components change the critical path significantly. In our evaluation, the time spent on AI-related and non AI-related components is 34.29 and 49.07 milliseconds for the average latency, 74.8 and 135.7 milliseconds for the 90th percentile latency, 152.2 and 466.5  milliseconds for the 99th percentile latency, except for the data preprocessing and communication latency, which indicates that an industry scale AI application benchmark suite is essential to depict the characteristics of a modern Internet service.

\textbf{The limitations for AI to ensure a good service.} The online inference module needs to load the trained model and conducts a forward computation to obtain the result. However, the depth or the size of a neural network model may largely affect the inference time. For comparison, we train a more complicated neural network for TensorFlow serving with the size of the model increasing from 184 MB to 253 MB. We find that the latency of TensorFlow serving increases sharply, with the average latency increasing from 30.78 to 125.71 milliseconds, and the 99th percentile latency increasing from 149.12 to 5335.12 milliseconds. Hence, the Internet service architects must perform a trade-off between the service quality and the depth or size of a neural network model. 

\textbf{The difference of micro-architectural behaviors.} We characterize the changes of micro-architectural behaviors from the following two aspects.

\begin{itemize}
\item \textbf{The difference between non AI-related and AI-related components.} We use perf to sample the cache behaviors of AI-related and non AI-related components. We find that comparing to the non AI-related components, the AI-related components suffer from more cache misses in memory hierarchy, especially L2 cache misses per Kilo instructions. TensorFlow serving suffers from 61 L2 cache misses per Kilo instructions, while the average number of non AI-related components is 37.

\item \textbf{The changes from a small neural network model to a large one.} We compare two neural network models for TensorFlow serving, with a smaller one 184 MB, and a larger one 253 MB. We also sample their cache behaviors. We find that with a larger model, the L3 cache misses per Kilo instructions increase extremely sharply, from 1.38 to 8.9, which incurs large memory backend stalls to fetch data from memory. And thus the 99th percentile latency increases more than thirty times.

\end{itemize}

\subsection{Evaluation of Offline Training}

%\begin{figure}
%\centering
%\subfigure[System Behavior with Different Patterns.]{
%\label{perf:a} %% label for first subfigure
%\includegraphics[scale=0.8]{figures/sm.eps}}
%\hspace{1in}
%\subfigure[Micro-architecture Behavior with Different Patterns.]{
%\label{perf:b} %% label for second subfigure
%\includegraphics[scale=0.8]{figures/ipc.eps}}
%\caption{Impact of Data Pattern on Data Motifs.}
%\label{perf} %% label for entire figure
%\end{figure}

\begin{figure}[tb]
\centering
\includegraphics*[scale=1.1]{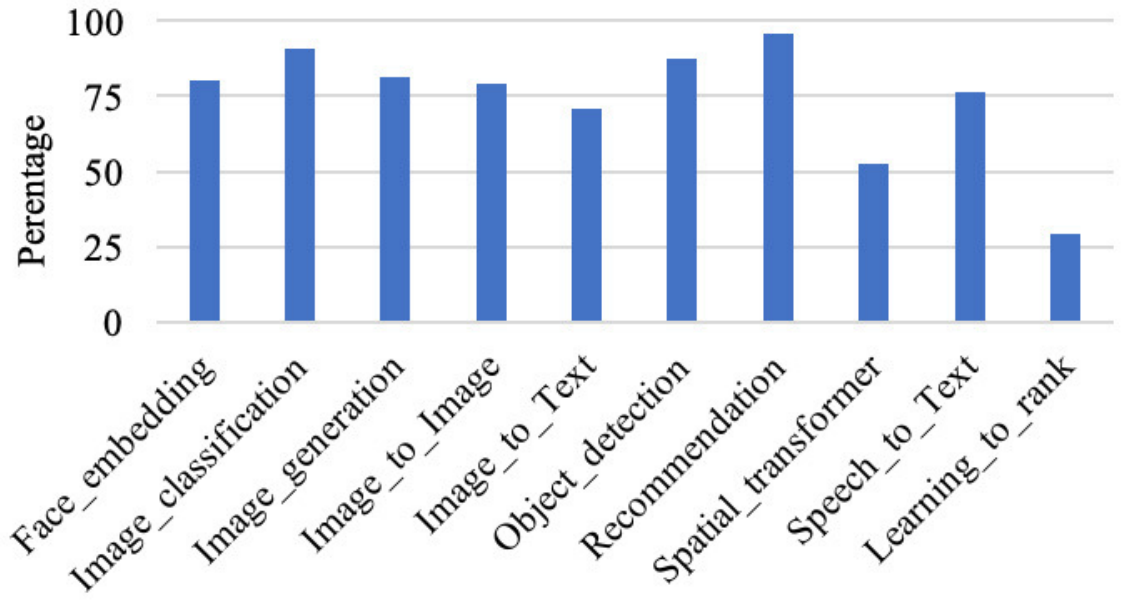}
\caption{SM Efficiency.} %\vspace{1pt}
\label{sm}
\end{figure}

GPU architecture contains multiple streaming multiprocessors (SM), each of which has a certain number of CUDA cores, memory registers, memory caches,  warp schedulers, etc. %Every CUDA core is an execute unit. 
The GPU efficiencies of running AI benchmarks are significantly important for both the GPU architecture design and AI system optimization. In this subsection, we mainly explore the GPU execution efficiency and evaluate the ten component benchmarks used in offline analyzer of the end-to-end AI application benchmark on Titan XP GPU. We choose the PyTorch implementations with the version of 1.1.0 for evaluation.

We comprehensively characterize the GPU efficiency, drilling down to functional-level running time breakdown and execution stall analysis. The overall execution performance of these ten component benchmarks are varying in terms of SM efficiency, which measures the percentage of time that the SM has one or more warps are active. Fig.~\ref{sm} shows the SM efficiency of each benchmark, with the value ranging from 29\% (Learning\_to\_rank) to 95\% (Recommendation). We find that some benchmarks like learning\_to\_rank have extremely low SM efficiency comparing to the other benchmarks. To discover the factors that impact the performance greatly, we first conduct running time breakdown analysis and decompose the benchmarks to the hotspot kernels or functions, then we find the GPU execution efficiency in terms of different percentage of stalls.
%and characterize their execution performance from functional level on GPUs.

\subsubsection{Running Time Breakdown}\label{timebreakdown}

\begin{figure}[tb]
\centering
\includegraphics*[scale=0.9]{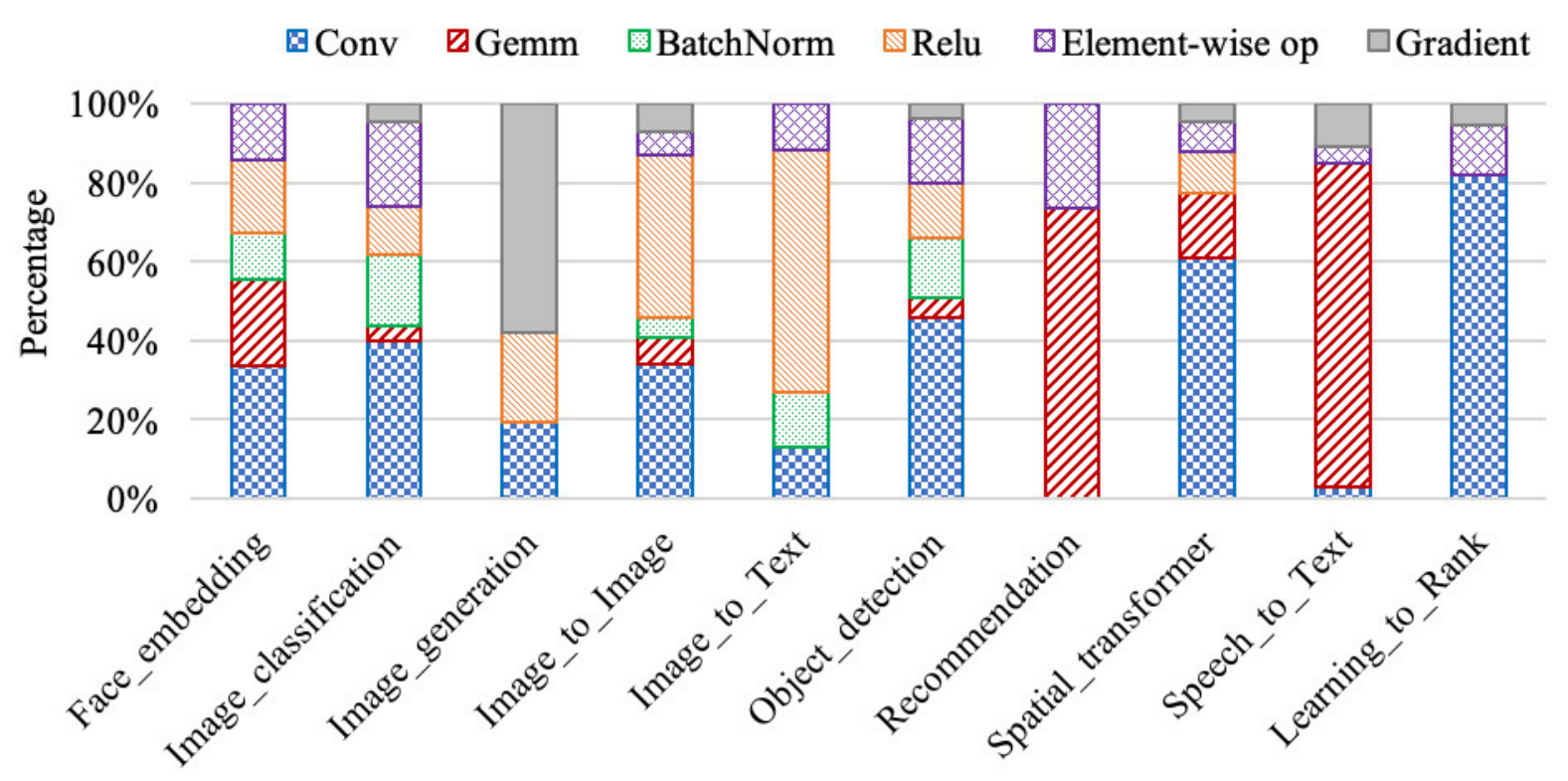}
\caption{Running Time Breakdown of the Ten Component Benchmarks.} %\vspace{1pt}
\label{time-breakdown}
\end{figure}

We use nvprof to trace the running time breakdown and find the hotspot functions that occupy more than 80\% of running time in total. 
Since each run involves  dozens of function calls, we single out the functions that occupy large proportions of running time and classify them into several categories of kernels according to their computation logics. 
Through statistics, we find that the most time-consuming functions among all the ten component benchmarks have much in common, and they are classified into six categories of kernels: convolution, general matrix multiply (gemm), batch normalization,  relu activation, element-wise operation and gradient calculation, which is consistent with our micro benchmarks and further indicates the decision of including them is correct. Note that each kernel contains a bunch of functions that solve the similar problem. For example, gemm kernel includes single or double precision floating general matrix multiply, etc. 
Fig.~\ref{time-breakdown} shows the running time breakdown of the above six kernels, using the average value of all involved functions within each kernel. Note that the remaining 20\% functions are not considered in this figure. 
Further, for each kernel, we summarize the typical functions that occupy a large proportion of running time among the ten component benchmarks, as shown in Table~\ref{func-sum}. We find that learning\_to\_rank spends too much time on the convolution category, with the mainstream function call of maxwell\_scudnn\_128x32\_stridedB\_splitK\_interior\_nn, whose SM efficiency is only 18.5\%, so this is the reason why leaning\_to\_rank has the lowest SM efficiency of 29\%.
We believe that the six kernels and these corresponding functions are the optimization directions not only for CUDA library optimizations but also for micro-architectural optimizations.  

\begin{table}[htbp]
%\scriptsize
\caption{Hotspot Functions for Each Kernel.}
\renewcommand\arraystretch{1.2}
\scriptsize
\label{func-sum}
\center %p{0.455in}|
\begin{tabular}{|p{1in}|p{4in}|}
\hline
\textbf{Kernel} & \textbf{Function Name} \\
\hline
\multirow{3}{*}{Convolution} &  maxwell\_scudnn\_128x128\_stridedB\_splitK\_interior\_nn \\
\cline{2-2}
& maxwell\_scudnn\_128x32\_stridedB\_splitK\_interior\_nn\\
\cline{2-2}
 & maxwell\_scudnn\_winograd\_128x128\_ldg1\_ldg4\_tile148n\_nt \\
\hline
\multirow{3}{*}{GEMM} & maxwell\_sgemm\_128x64\_nt \\
\cline{2-2}
& maxwell\_sgemm\_128x64\_nn \\
\cline{2-2}
& sgemm\_32x32x32\_NN\_vec \\
\hline
\multirow{3}{*}{BatchNorm} &  cudnn::detail::bn\_fw\_tr\_1C11\_kernel\_NCHW  \\
\cline{2-2}
& cudnn::detail::bn\_bw\_1C11\_kernel\_new  \\
\cline{2-2}
& batch\_norm\_backward\_kernel\\
\cline{2-2}
& at::native::batch\_norm\_backward\_kernel \\
\hline
\multirow{2}{*}{Relu} & maxwell\_scudnn\_128x128\_relu\_small\_nn  \\
\cline{2-2}
& maxwell\_scudnn\_128x128\_relu\_small\_nn  \\
\cline{2-2}
& maxwell\_scudnn\_128x32\_relu\_interior\_nn \\
\hline
\multirow{2}{*}{\tabincell{l}{Element-wise}} & element-wise add kernel \\
\cline{2-2}
& element-wise threshold kernel \\
\hline
\multirow{2}{*}{Gradient} & cudnn::detail::dgrad\_engine \\
\cline{2-2}
& cudnn::detail::dgrad\_alg1\_engine \\
\hline

\end{tabular}
\end{table}

\subsubsection{GPU Execution Efficiency Analysis}

Focusing on the above six most time-consuming kernels, we evaluate the stalls of these kernels, including instruction fetch stall (Inst\_fetch), which indicates the percentage of stalls because the next assembly instruction has not yet been fetched, execution dependency stall (Exe\_depend), which is the percentage of stalls because an input required by the instruction is not yet available, memory dependency stall (Mem\_depend), which is the percentage of stalls because a memory operation cannot be performed due to the required resources not being available or fully utilized, texture stall (Texture), which is the percentage of stalls because of the under-utilization of the texture sub-system,  synchronization stall (Sync), which is the percentage of stalls due to a syncthreads call, constant memory dependency stall (Const\_mem\_depend), which is the percentage of stalls because of immediate constant cache miss, pipe busy stall (Pipi\_busy), which is percentage of stalls because a compute operation cannot be performed because the compute pipeline is busy, and memory throttle stall (Mem\_throttle), which is the percentage of stalls due to large pending memory operations~\cite{nvprof}.

\begin{figure}[tb]
\centering
\includegraphics[scale=1]{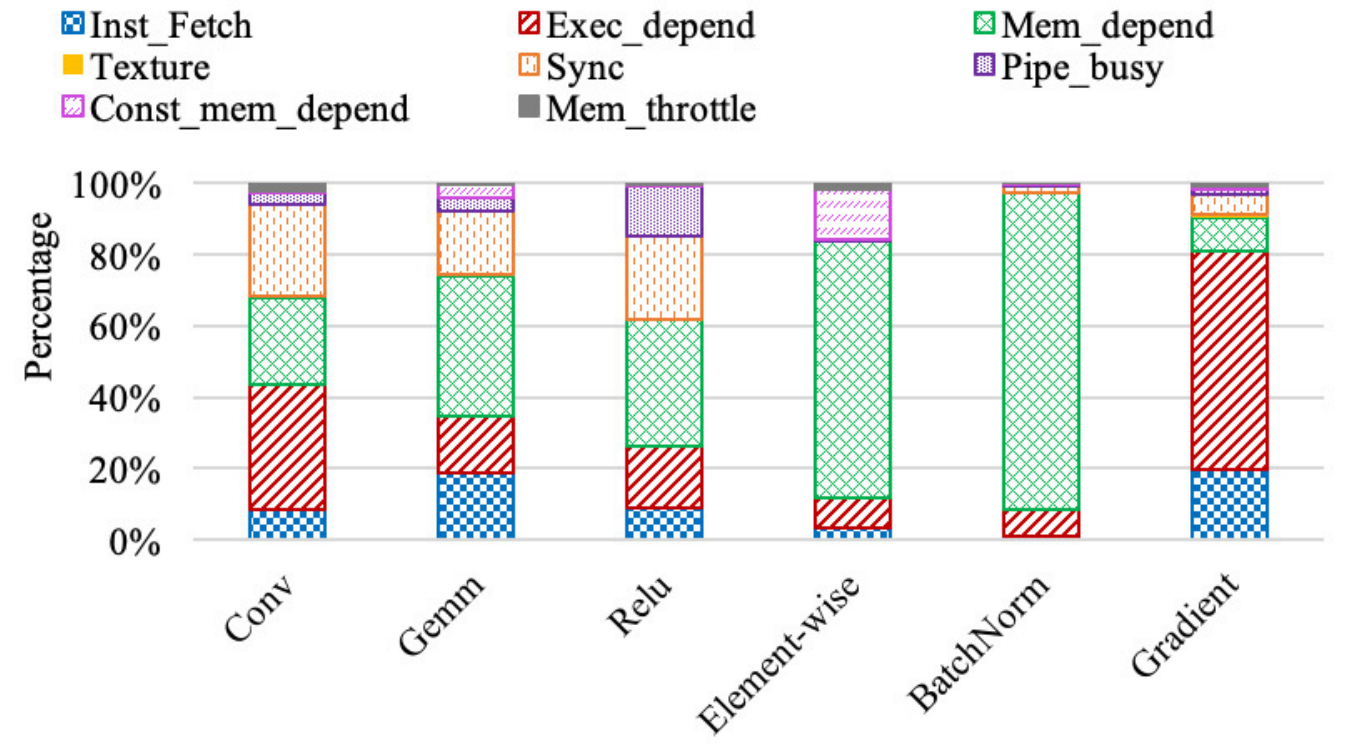}
\caption{Stall Breakdown of Each Kernel.} %\vspace{1pt}
\label{stall-breakdown}
\end{figure}

%In order to find the micro-architectural bottleneck, we further do the stall analysis at the functional level. 
The breakdown of eight kinds of stalls of each kernel is shown in Fig.~\ref{stall-breakdown}.
We find that the top two GPU execution stalls are memory dependency stalls, and execution dependency stalls. For example, for Element-Wise kernels, the memory dependency stalls occupy a very large proportion of 68\%, thus resulting low SM efficiency of about 50\%.
The memory dependency stalls may occurs due to the high cache misses and thus the load/store resources are not available. The optimization strategies include optimizing date alignment, data locality, and data access pattern. The execution dependency stalls may occur due to the low instruction-level parallelism, so exploiting ILP may alleviate partial execution dependency stalls to a certain degree.

We also identify the functional level stalls, including the hotspot functions 
 illustrated in Table~\ref{func-sum}, to provide potential optimization guidelines for function calls. We find that performing analytics of stalls on a function level in addition to a kernel level is helpful. For example, the memory dependency stall percentage of maxwell\_scudnn\_128x32\_stridedB\_splitK\_interior\_nn function in ``convolution" category achieves 61\%, however, the percentage of maxwell\_sgemm\_128x64\_nn function in ``GEMM" category is 18\%, indicating that different optimization strategies are needed to achieve maximum performance improvement.

\section{Conclusion}

This paper presents the first industry standard Internet service AI benchmark suite with seventeen industry partners.  
We propose and implement a highly extensible, configurable, and flexible AI benchmark framework, and identify sixteen prominent AI problem domains from three most important Internet services domains: search engine, social network, and e-commerce.
%, and implement component benchmarks targeting
%those domains accordingly.  Also, we profile 12 most intensive unites of computation accross those components benchmarks as micro benchmarks. 
On the basis of the AIBench framework, we
design and implement the first end-to-end Internet service AI benchmark suite, with an
underlying e-commerce searching model. %Our AI benchmark framework is also extensible for other industry applications. 
 On the CPU and GPU clusters, we perform a preliminary evaluation of the end-to-end application benchmark. We observe that the AI-related components significantly change the
critical paths and workload characterization of  the Internet service, which justifies the end-to-end AI application benchmark. % The specification, source code, and performance numbers are publicly available from the benchmark council web site （deleted for double-blind review）.

%end-to-end datacenter AI benchmark framework and suite---AIBench, . Through the sixteen problem domains identified from internet service applications, we design our easily configured and extended framework with essential loosely coupled modules. These modules collectively constitute an end-to-end application benchmarks, while individually contain a series of micro and component benchmarks. Based on the framework, we construct AIBench modelling an intelligent e-commerce search scenario, equipped with primary modules for online serving and offline analytics.  Overall, AIBench provides an end-to-end e-commerce AI benchmark, 16 component benchmarks, 12 micro benchmarks, and a highly configurable and flexible framework that can be extended to the other internet service applications.

%\bibliographystyle{ieeetr}
%\bibliography{sample-bibliography}

\end{document}